\documentclass{article}

\usepackage{arxiv}
\usepackage[utf8]{inputenc} 
\usepackage[T1]{fontenc}    
\usepackage{hyperref}       
\usepackage{url}            
\usepackage{booktabs}       
\usepackage{amsfonts}       
\usepackage{nicefrac}       
\usepackage{microtype}      
\usepackage{cleveref}       
\usepackage{lipsum}         
\usepackage{graphicx}
\usepackage{natbib}
\usepackage{doi}
\usepackage{multirow}
\usepackage{url}
\usepackage{CJKutf8}

\title{StyleBERT: Chinese Pretraining by Font Style Information}

\author{Chao Lv \footnotemark[1] , Han Zhang \footnotemark[1] , XinKai Du \footnotemark[2] , Yunhao Zhang, Ying Huang, Wenhao Li, Jia Han, Shanshan Gu\\
	Science \& Technology Center, Sunshine Insurance Group Co., Ltd\\
	Beijing, China \\
	\texttt{\{lvchao, zhanghan03, duxinkai, zhangyunhao, huangying,}\\
	\texttt{liwenhao01, hanjia,  gushanshan\}-ghq@sinosig.com} \\
}

\renewcommand{\shorttitle}{StyleBERT}

\hypersetup{
pdftitle={A template for the arxiv style},
pdfsubject={q-bio.NC, q-bio.QM},
pdfauthor={Chao Lv, Han Zhang, Xinkai Du and etc..},
pdfkeywords={pre-trained language model, Chinese},
}

\begin{document}

\maketitle

\renewcommand{\thefootnote}{\fnsymbol{footnote}}
\footnotetext[1]{These authors contributed equally to this work and should be considered co-first authors.}
\footnotetext[2]{Corresponding author.}

\begin{abstract}
	With the success of down streaming task using English pre-trained language model, the pre-trained Chinese language model is also necessary to get a better performance of Chinese NLP task. Unlike the English language, Chinese has its special characters such as glyph information. So in this article, we propose the Chinese pre-trained language model StyleBERT which incorporate the following embedding information to enhance the savvy of language model, such as word, pinyin, five stroke and chaizi. The experiments show that the model achieves well performances on a wide range of Chinese NLP tasks.
\end{abstract}


\section{Introduction}

Large scale pre-trained models such as BERT \cite{devlin2019bert} have been widely used to improve various downstream tasks such as text classification \cite{reimers2019sentencebert, sun2020finetune}, reading comprehension \cite{xu2019bert, ramnath2020interpreting}, summarization \cite{liu2019finetune, liu2019text}, question answering \cite{wang-etal-2019-multi, qu2019bertqa} and etc.. The pre-trained models are also proved to be effective for various Chinese NLP tasks \cite{yang2019bert, jia-etal-2020-entity}. In \cite{bengio2014representation}, the authors propose that a expressive language model should not only capture the implicit linguistic rules but also the common sense of knowledge hiding in the text data, such as lexical meanings, syntactic structures, semantic roles, and even pragmatics. 

Chinese is a kind of logographic language which is distinguished from English or German \cite{dai-cai-2017-glyph}. To be better grasp the \textit{``common sense of knowledge''} for Chinese language, a lot of pre-trained language model for Chinese processing have been proposed. A lot of works have incorporated Chinese glyph information into neural models but not as large-scale pretraining \cite{sun2014radicalenhanced, liu2017learning, tao2019chinese, meng2020glyce}. And also in \cite{dai-cai-2017-glyph}, the authors introduce the spatio-structural patterns of Chinese glyphs which are rendered in raw pixels as a novel glyph-aware embedding of Chinese characters. The glyph information has proved to be effective for the context of two basic Chinese NLP tasks of language modeling and word segmentation.

\begin{CJK*}{UTF8}{gbsn}
Except glyph information, the pinyin information is also considered to be constructed the pre-trained Chinese language model \cite{sun-etal-2021-chinesebert}. For the Chinese character, there are some cases that the same character have multiple pronunciations and each pronunciation is associated with a idiographic meaning which also correspond with a specific pinyin declaration. For example, the Chinese character ``数'' has three different pronunciations. The first is ``shǔ'' which means ``count'' and it is a verb. The second is ``shù'' which means ``number'' and it is a noun. The third is ``shuò'' which means ``frequently'' and it is an adverb. So we can see that the same Chinese character ``数'' is expressed the different pronunciation, different meaning and even different part of speech at semantic level and the syntax level.  
\end{CJK*}

In this article, we propose StyleBERT which first introduce the ``chaizi'' information into the Chinese pre-training process which is proved to be more expressive than the other Chinese glyph information such as different font style information \cite{sun-etal-2021-chinesebert}, five-stroke information \cite{li2021mfener}. To be better grasp the semantic information from the raw Chinese characters, word, pinyin, five-stroke and chaizi information are also put into together. In next section, we will describe the StyleBERT model in detail.

\section{Related Works}

In this section, we revisit the techniques of the representative Chinese pre-trained language models in the recent natural language processing filed. And the key components of these models are elaborated in the following subsections.


\subsection{BERT-wwm}

In the original BERT, a WordPiece tokenizer \cite{wu2016googles} was used to split the text into Word Piece tokens, where some words will be split into several small fragments \cite{BERT-wwm}. However in \cite{BERT-wwm}, the authors proposed the whole word masking (wwm) method to mitigate the drawback of masking only a part of the whole word, which make it is easier for the model to predict. In Chinese, the character is the smallest semantics element, and mostly we use the Chinese word to present the semantic in contemporary Chinese language. So in this place we do not need to split the word into small fragments, for the case that Chinese characters are not formed by alphabet-like symbols. So we can use the traditional Chinese Word Segmentation (CWS) tool to split the text into words and adopt whole word masking in Chinese to mask the word instead of individual Chinese characters. An example of the whole word masking is depicted in Table \ref{tabmacmask}.


\begin{table}[]
\label{tabmacmask}
\centering
\tiny
\begin{CJK*}{UTF8}{gbsn}
\begin{tabular}{@{}lll@{}}
\toprule
 & Chinese &  English  \\ \midrule
 Original Sentence &   使\ 用\ 语\ 言\ 模\ 型\ 来\ 预\ 测\ 下\ 一\ 个\ 词\ 的\ 概\ 率\ 。  &  we use a language model to predict the probability of the next word. \\
 + CWS &  使用\ 语言\ 模型\ 来\ 预测\ 下\ 一个\ 词\ 的\ 概率\ 。  & -  \\
 + BERT Tokenizer &  使\ 用\ 语\ 言\ 模\ 型\ 来\ 预\ 测\ 下\ 一\ 个\ 词\ 的\ 概\ 率\ 。       &  we use a language model to pre \#\#di \#\#ct the pro \#\#ba \#\#bility of the next word .  \\ \midrule
 Original Masking &   使\ 用\ 语\ 言\ [M]\ 型\ 来\ [M]\ 测\ 下\ 一\ 个\ 词\ 的\ 概\ 率\ 。      &  we use a language [M] to [M] \#\#di \#\#ct the pro [M] \#\#bility of the next word . \\
 + WWM  & 使\ 用\ 语\ 言\ [M]\ [M]\ 来\ [M]\ [M]\ 下\ 一\ 个\ 词\ 的\ 概\ 率\ 。  & we use a language [M] to [M] [M] [M] the [M] [M] [M] of the next word . \\ 
 ++ N-gram Masking  &  使\ 用\ [M]\ [M]\ [M]\ [M]\ 来\ [M]\ [M]\ 下\ 一\ 个\ 词\ 的\ 概\ 率\ 。 &  we use a [M] [M] to [M] [M] [M] the [M] [M] [M] [M] [M] next word . \\
 +++ Mac Masking   &  使\ 用\ 语\ 法\ 建\ 模\ 来\ 预\ 见\ 下\ 一\ 个\ 词\ 的\ 几\ 率\ 。 & we use a text system to ca \#\#lc \#\#ulate the po \#\#si \#\#bility of the next word . \\
 \bottomrule
\end{tabular}
\end{CJK*}
\caption{Different masking strategies \cite{MacBER}}
\end{table}

\subsection{MacBERT}

For the model of \cite{MacBER}, the authors modify the MLM task as following.

\begin{itemize}
    \item The N-gram masking are used as the whole word masking strategies for selecting candidate tokens for masking. And also the word-level unigram is changed to 4-gram with a percentage of 40\%, 30\%, 20\%, 10\%.
    \item The similar words which are obtained by using Synonyms toolkit \cite{Synonyms:hain2017} are proposed as the masking purpose instead of masking with \texttt{[MASK]} for the case that these masking tokens never appear in the fine-tune stage. And if we choose an N-gram to mask, the similar words will be found individually. While there is no similar word, ``we will degrade to use random word replacement'' \cite{MacBER}. And this will happen rarelly.
    \item The 15\% percentage of input words are used for masking, and 80\% of this part will be replaced with similar words, 10\% part will be replaced with a random word, and the rest of 10\% keep as the original.
\end{itemize}

\subsection{Chinese-BERT}

In \cite{sun-etal-2021-chinesebert}, Chinese-BERT is proposed which incorporates the \emph{glyph} and \emph{pinyin} information of Chinese characters into language pretraining. For capturing character semantics from the visual features, the authors build the glyph embedding which is based on different fonts of a Chinese character. For handling the highly prevalent heteronym phenomenon in Chinese (the same character has different pronunciations with different meanings), the pinyin embedding is also constructed to characterize the pronunciation of Chinese characters. So we can model the distinctive semantic property of a Chinese character by a fusion embedding which is combined by the glyph embedding, the pinyin embedding and the character embedding.

\begin{figure}[h]
\centering
\includegraphics[width = 0.7\hsize]{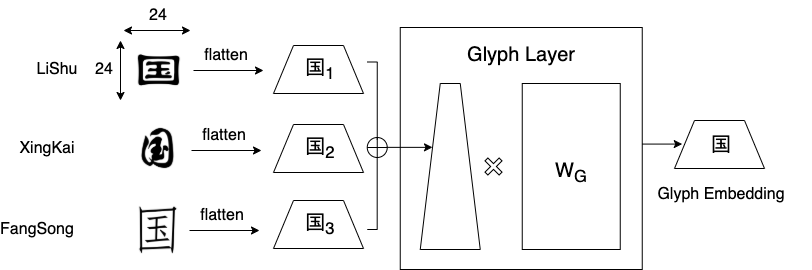}
\caption{ChineseBERT Glyph Embedding}
\label{fig:chinesebert-glyph}
\end{figure}

In ChineseBERT, the glyph embedding is constructed by using three types of Chinese fonts – LiShu, XingKai and FangSong which is followed from this article \cite{meng2020glyce}. And each font character can be instantiated as a $24 \times 24$ image with floating point pixels ranging from 0 to 255. The $24 \times 24 \times 3$ vector is first flattened to a 2,352 vector. The flattened vector is fed to an FC layer to obtain the output glyph vector which is shown in Figure \ref{fig:chinesebert-glyph}.

\section{StyleBERT}

\begin{figure}[h]
\centering
\includegraphics[width = 0.7\hsize]{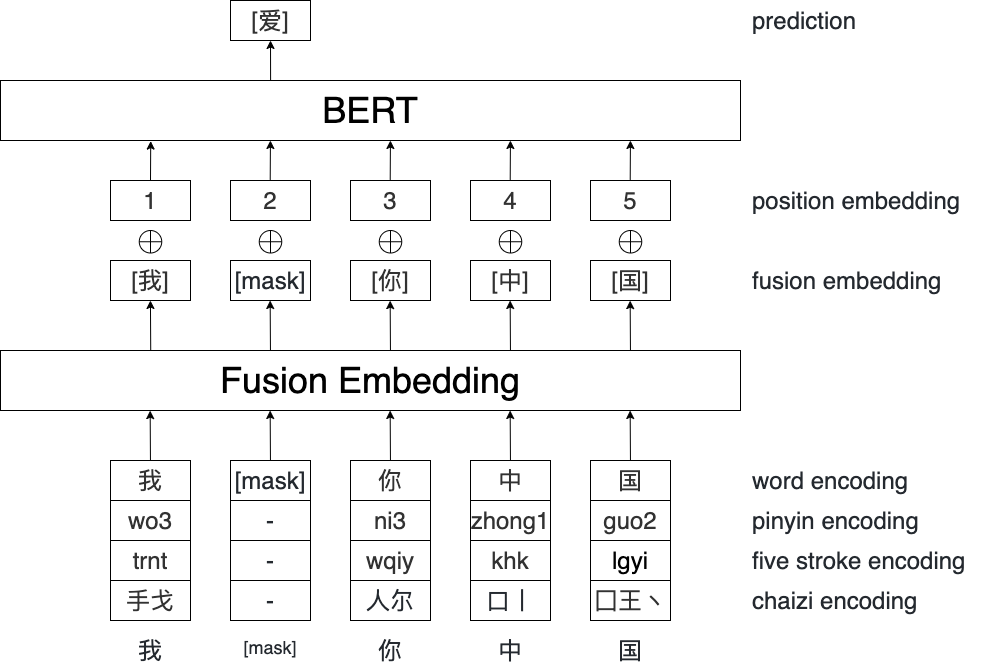}
\caption{The general workflow of StyleBERT model. Four embedding vectors of each token is fed into Fusion Layer for element wise addition followed by a fully connected layer to generate the fusion embedding vector.}
\label{fig:Stylebert}
\end{figure}

\subsection{Overview}
As shown in Figure \ref{fig:fusion}, each input Chinese character is embedded into word vector, pinyin vector, five stroke vector and chaizi vector. These embedding vectors are sent into the Fusion layer to merge into the fusion vector. The fusion embedding is used to replace token embedding vector in original BERT-base model.

\subsection{Input}

\subsubsection{Feature Encoding}

\begin{figure}[h]
\centering
\includegraphics[width = 0.7\hsize]{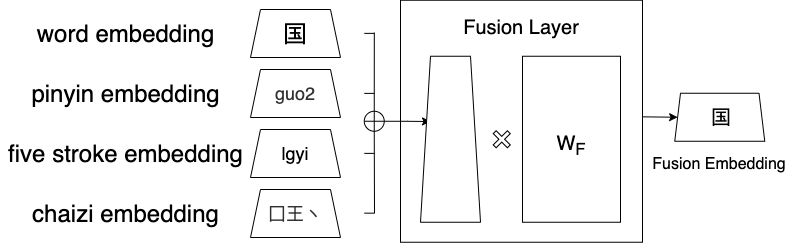}
\caption{An example of character embedding. As shown in the figure, the given Chinese character is embedded into four different vectors: word embedding, pinyin embedding, five stroke embedding and chaizi embedding. Pinyin embedding vector carry's phonetic information; Glyph features are represented by both five stroke embedding and chaizi embedding. }
\label{fig:fusion}
\end{figure}

\textbf{Phoneme Feature Embedding}: Compared with English content, whose phonetic features of tokens are highly dependent on the sequential formation of each token, in Chinese contents, each character's phonetic feature can hardly be related to its word encoding vector, since each Chinese character is basically a single picture without any sequential order. During pinyin encoding, each given token is mapped to a 8-dimension vector. Each pinyin vector is composed of two parts. For instance, in Figure \ref{fig:fusion}, the Chinese character's pinyin encoding is [`g',`u',`o',`2']. The first part [`g',`u',`o'] is the Latin encoding of Chinese character's pronunciation; the second part [`2'] reflects the tone of the character. The encoding is then padded to length of 8 and sent into a embedding layer and a fusion layer to form the final pinyin embedding vector.

\textbf{Glyph Feature Embedding}: Glyph feature also plays an important role during Chinese language modeling. Each Chinese character is composed of one or more sub-characters or radicals. Unlike English words, the radicals in Chinese character are not combined strictly in horizontal format. The combination of Chinese radicals can also be vertical, surrounding, half-surrounding, etc, which makes the structure of Chinese characters even harder to be interpreted. Therefore, five stroke encoding and chaizi encoding are applied to encode the radical structure of Chinese characters. 

\begin{figure}[h]
\centering
\includegraphics[width = 0.5\hsize]{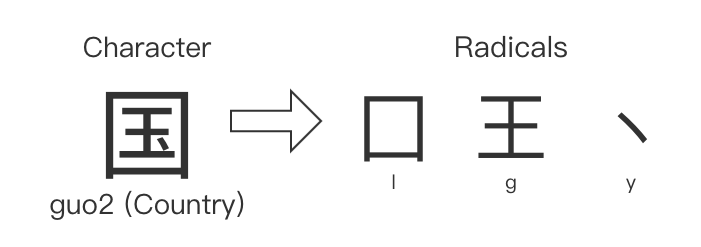}
\caption{Five Stroke encode example.}
\label{fig:fivestroke}
\end{figure}

\begin{CJK*}{UTF8}{gbsn}
For five stroke encoding, all Chinese radicals are mapped into 26 English characters. The interpretation rule follows left-to-right, above-to-below and outer-to-inner orders. As shown in Figure \ref{fig:fivestroke}, the Chinese character `国', which means country in English, can be split into three radicals: `囗', `王' and `丶'. In five stroke encoding, these three radicals are mapped to three English characters `l', `g' and `y'. The total volume of Chinese radical is about 1.6k. 
five stroke encoding greatly reduces feature space size and captures geometrical structure of each character. There are existing works using five stroke encoding for Chinese language modeling. However, one noticeable drawback of five stroke is that it over reduces the feature space, which makes the model harder to converge. To reduce the impact of this problem, chaizi encoding is introduced to the model. 
\end{CJK*}

Unlike five stroke, chaizi encoding assigns different indies to each Chinese radical, which largely increased the encoding space and maintained more information. In this project, a open source github repository \cite{chaizi} is referenced during chaizi encoding, which contains radical combinations for each Chinese characters.

\begin{figure}[h]
\centering
\includegraphics[width = 0.45\hsize]{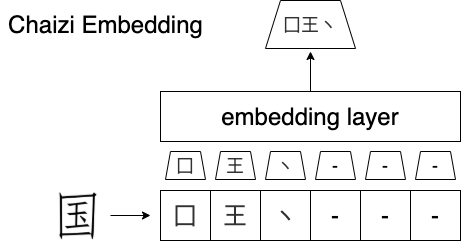}
\caption{Chaizi embedding example.}
\label{fig:chaizi}
\end{figure}

After generating Pinyin, five stroke and Chaizi encoding vectors, these vectors are embedded into embedding vectors. In this paper, multiple embedding approaches are tested:

\begin{itemize}
    \item  \textbf{TextCNN}: The first embedding approach is Text-CNN. Feature encoding vectors are sent into a Text-CNN layer which uses multiple filters to capture encoding features. This approach performs well in many downstream tasks. However, one drawback is that the Text-CNN layer takes too much time during training stage, especially for longer corpus.
    
    \item \textbf{RNN with Attention}: To accelerate the training process, the Text-CNN layer is replaced by a RNN layer followed by an attention layer, which is introduced in \cite{PengZhou2016AttentionBasedBL}. In addition, the team added a skip connection which adds the inputs encoding vector to the output vector of RNN layer. This approach not only generates good results but also costs less time. 
\end{itemize}

\begin{CJK*}{UTF8}{gbsn}
As an chaizi embedding example, in Figure \ref{fig:chaizi}, the Chinese character `国' is first encoded into vector [`囗', `王', `丶'] and padded to max length. Then, each different radial is mapped to a unique induce before sending into embedding layer. Pinyin and five stroke embedding follows the similar process as chaizi embedding.
\end{CJK*}

These embedding vectors are then sent into the Fusion Layer which integrates all feature embedding vectors into a single Fusion vector by concatenation and a linear layer, as shown in Figure \ref{fig:fusion}.

\section{Experimental Setups}

\subsection{Data Processing}
In this paper, we selected the Chinese corpus from CLUE organization, CLUECorpus2020, a large-scale corpus that contains four sub-corpora. We extracted 80M texts from the three sub-corpora of news, Wikipedia and comments for pretraining, containing a total of 3B Chinese characters.

\subsection{Setups for Pre-Trained Language Models}

\begin{table}[htbp]
    \centering    
    \begin{tabular}{lcccc}
    \toprule
        & \textbf{BERT} & \textbf{BERT-wwm} & \textbf{MacBERT} & \textbf{StyleBERT} \\
    \midrule
    \textbf{Data Source} & Wikipedia & Wikipedia & Heterogeneous & CLUECorpusSmall \\
    \textbf{Vocab Size} & 21K   & 21K   & 21K   & 21K \\
    \textbf{Input Unit} & Char  & Char  & Char  & Char \\
    \textbf{Masking} & CM    & WWM   & WWM/N & CM \\
    \textbf{Task} & MLM/NSP & MLM   & MAC/SOP & MLM \\
    \textbf{Hidden Activation} & GeLU  & GeLU  & GeLU  & GeLU \\
    \textbf{Optimizer} & AdamW & LAMB  & LAMB  & AdamW \\
    \textbf{Training Steps} & –-     & 2M    & 1M    & 1.6M \\
    \textbf{Init Checkpoint} & random & BERT  & random & BERT \\
    \textbf{\# Token} & 0.4B  & 5.4B  & 5.4B  & 3B \\
    \bottomrule
    \end{tabular}%
    \setlength{\abovecaptionskip}{0.2cm}  
    \setlength{\belowcaptionskip}{-0.2cm} 
    \caption{Training details of Chinese pre-trained language models. CM: Char Masking, WWM: Whole Word Masking, N: N-gram, MLM: Masked Language Model, NSP: Next Sentence Prediction, MAC: MLM-As-Correlation. SOP: Sentence Order Prediction.}
    \label{tab:Model}%
  \end{table}%

We used the masking strategy of Char Masking(CM) for StyleBERT. \cite{devlin2019bert} indicated that it can alleviate the out of vocabulary problem. And we considered that this method is an easy-to-implement word masking approach that doesn't require word segment and relies on external text resources. Beside, in order to learn the semantic and syntactic information from the context in a better way, instead of training the basic model from scratch, we chose to continue training with BERT's vocabulary and weight. The detail of the training process is shown in Table \ref{tab:Model}. StyleBERT selected the original ADAM and weight decay optimizer for optimization. We trained the model 500K steps with a batch size of 512, an initial learning rate of 1e-4. After pre-training, we chose the same way as BERT (\cite{devlin2019bert}) to complete the fine-tuning of downstream tasks by this pre-trained model.
\subsection{Setups for Fine-tuning Tasks}

\begin{table}[h]
    \centering    
    \begin{tabular}{cl|ccc|ccc}
    \toprule
    \textbf{Task} & \textbf{Dataset} & \textbf{Train \#} & \textbf{Dev \#} & \textbf{Test \#} & \textbf{MaxLen} & \textbf{Batch} & \textbf{InitLR} \\
    \midrule
    \textbf{MRC} & CMRC 2018 & 14k(10k) & 4.5k(3.2k) & 1.4k(1k) & 512   & 16    & 3e-5 \\
    \midrule
    \textbf{NLI} & XNLI  & 392K  & 2.5K  & 5K    & 64    & 512   & 8e-5 \\
    \midrule
    \multirow{2}[2]{*}{\textbf{TC}} & THUCNews & 50K   & 5K    & 10K   & 512   & 8     & 2e-5 \\
        & ChnSentiCorp & 9.6K  & 1.2K  & 1.2K  & 512   & 8     & 2e-5 \\
    \midrule
    \multirow{2}[2]{*}{\textbf{SPM}} & LCQMC & 240K  & 8.8K  & 12.5K & 32    & 256   & 2e-5 \\
        & BQ Corpus & 100K  & 10K   & 10K   & 32    & 64    & 2e-5 \\
    \midrule
    \multirow{2}[2]{*}{\textbf{NER}} & Weibo & 1.3K  & 4.3K  & 4.3K  & 275   & 2     & 3e-5 \\
        & OntoNotes & 15K   & 270  & 270  & 275   & 2     & 3e-5 \\
    \bottomrule
    \end{tabular}%
    \setlength{\abovecaptionskip}{0.2cm}  
    \setlength{\belowcaptionskip}{-0.2cm} 
    \caption{ Data statistics and hyper-parameter settings for different ﬁne-tuning tasks.}
    \label{tab:settings}%
  \end{table}%

We conducted extensive experiments on various natural language processing tasks to ensure a more comprehensive testing of these pre-trained language models. Details of the tasks are shown in Table \ref{tab:settings}. Specifically, we selected the following five popular tasks with datasets covering a wide range of text lengths, i.e., from sentence level to document level.

\begin{itemize}
\item Machine Reading Comprehension (MRC): CMRC 2018 (\cite{mrc})
\item Natural Language Inference (NLI): XNLI (\cite{xnli})
\item Text Classification (TC): THUCNews (\cite{THUCNews}), ChnSentiCorp (\cite{ChnSentiCorp})
\item Sentence Pair Matching (SPM): LCQMC (\cite{lcqmc}), BQ Corpus (\cite{bq})
\item Named Entity Recognition: Weibo (\cite{Weibo}), OntoNotes (\cite{Ontonotes})
\end{itemize}

Since the pre-training data of various Chinese pre-training language models vary relatively extensive, such as ERNIE (\cite{sun2019erniea}), ERNIE 2.0 (\cite{sun2019ernieb}). In order to make a relatively fair comparison between different models, StyleBERT only compares with BERT (\cite{devlin2019bert}), BERT-wwm (\cite{BERT-wwm}), BERT-wwm-ext, RoBERTa (\cite{YinhanLiu2019RoBERTaAR}) and MacBERT (\cite{MacBER}). Meanwhile, the same hyperparameters (e.g., maximum length, warm-up steps, etc.) were used for each data set.

\section{Results}

\subsection{Machine Reading Comprehension(MRC)}
\begin{table}[htbp]
  \centering
    \begin{tabular}{lcc}
    \toprule
          & \multicolumn{2}{c}{\textit{CMRC}} \\
    \textbf{Model} & \textbf{Dev} & \textbf{Test} \\
    \midrule
    BERT  & 63.4 & 31.1 \\
    BERT-wwm & 64.0 & 29.8 \\
    BERT-wwm-ext & 64.3 & 30.5 \\
    RoBERTa & 67.3 & 31.1 \\
    MacBERT & \textbf{67.1} & \textbf{33.4} \\
    StyleBERT & 65.7 & 31.8 \\
    \bottomrule
    \end{tabular}%
       \setlength{\abovecaptionskip}{0.2cm}  
  \setlength{\belowcaptionskip}{-0.2cm} 
  \caption{Performance of StyleBERT and other models on the CMRC 2018 dataset.EM is reported for comparison.}
  \label{tab:addlabel}%
\end{table}%

MRC is a task which answers questions based on a given context. We conducted experiments on the CMRC 2018 dataset. CMRC 2018 is for Chinese Machine Reading Comprehension especially which is a span-extraction style dataset. 

In this work, we reproduced the results of BERT-base, BERT-wwm and BERT-wwm-ext on this dataset by using a open-source text-classification script $run\_qa\_no\_trainer.py$ \footnote{https://github.com/huggingface/transformers/blob/master/examples/pytorch/question-answering/run\_qa\_no\_trainer.py} from Huggingface's Transformers library. The results are showed in Table 3. Compared to the results published on their original paper, we got slightly lower scores in our reproduction for these three models. As we can see, except MacBERT, styleBERT achieved the best performance on the validation and test dataset of CMRC 2018.

\subsection{Natural Language Inference (NLI)}
\begin{table}[htbp]
  \centering
    \begin{tabular}{lcc}
    \toprule
          & \multicolumn{2}{c}{\textit{XNLI}} \\
    \textbf{Model} & \textbf{Dev} & \textbf{Test} \\
    \midrule
    BERT  & 77.8  & 77.8 \\
    BERT-wwm & 79.0    & 78.2 \\
    BERT-wwm-ext & 79.4  & 78.7 \\
    RoBERTa & 80 & 78.8 \\
    MacBERT & \textbf{80.3}  & \textbf{79.3} \\
    StyleBERT & 79.4  & 78.8 \\
    \bottomrule
    \end{tabular}%
  \setlength{\abovecaptionskip}{0.2cm}  
  \setlength{\belowcaptionskip}{-0.2cm} 
  \caption{Performance of StyleBERT and other models on the XNLI dataset.Accuracy is reported for comparison.}
  \label{tab:NLI_result}%
\end{table}%

NLI can reflect the model's understanding of semantics. Models will be used to determine the entailment relationship between ``Premis'' and ``Hypothesis''. We chose Cross-lingual Natural Language Inference (XNLI) dataset was used to train and evaluate the performance of the models. The XNLI dataset contains 390K training sets, 2.5K dev sets, and 5K test sets. In this task, the model was fed two sentences and return an answer that may be ``Entailment'', ``Contradiction'' or ``Neutral''.  \par
The text of XNLI dataset is mixed in Chinese and English, while StyleBERT focuses more on Chinese text. As shown in  Table \ref{tab:NLI_result}, StyleBERT outperforms most models.

\subsection{Text Classification (TC)}

\begin{table}[htbp]
  \centering
    \begin{tabular}{lcccc}
    \toprule
          & \multicolumn{2}{c}{\textit{THUCNews}} & \multicolumn{2}{c}{\textit{ChnSentiCorp}} \\
    \textbf{Model} & \multicolumn{1}{c}{\textbf{Dev}} & \multicolumn{1}{c}{\textbf{Test}} & \multicolumn{1}{c}{\textbf{Dev}} & \multicolumn{1}{c}{\textbf{Test}} \\
    \midrule
    BERT  & 97.7  & 97.8  & 95.2  & 95.3  \\
    BERT-wwm & 98.0  & 97.8  & 95.1  & 95.4  \\
    BERT-wwm-ext & 97.7  & 97.7  & 95.4  & 95.3  \\
    RoBERTa & 98.2 & 97.8 & 94.9 & 95.6 \\
    MacBERT & \textbf{98.2}  & 97.7  & 95.2  & 95.6  \\
    StyleBERT & 97.9  & \textbf{97.8}  & \textbf{95.7}  & \textbf{96.0}  \\
    \bottomrule
    \end{tabular}%
  \setlength{\abovecaptionskip}{0.2cm}  
  \setlength{\belowcaptionskip}{-0.2cm} 
  \caption{Performances of StyleBERT and other models on TC datasets THUCNewsandTNEWS and ChnSentiCorp.Accuracy is reported for comparison.}
  \label{tab:TC_result}%
\end{table}%

TC is a task that assigns a sentence or document into one or more categories within a given classification system. In our work, we did the evaluation on this task by using two datasets: THUCNews and ChnSentiCorp. THUCNews is a dataset containing news from different domains. It includes 50K news in total covering 10 fields (evenly distributed) including finance, technology, etc. ChnSentiCorp is a dataset that contains 9.6K hotel comments. And it is often used for a binary emotional classification task which determines a comment is positive or negative. In our results, it showed that StyleBERT partly improved the baseline of THUCNews and ChnSentiCorp, as the baseline already had a very high accuracy on these two data sets .

\subsection{Sentence Pair Matching (SPM)}
\begin{table}[htbp]
  \centering
    \begin{tabular}{lcccc}
    \toprule
          & \multicolumn{2}{c}{\textit{BQ}} & \multicolumn{2}{c}{\textit{LCQMC}} \\
    \textbf{Model} & \textbf{Dev} & \textbf{Test} & \textbf{Dev} & \textbf{Test} \\
    \midrule
    BERT  & 85.4  & 85.2  & 89.4  & 87.3  \\
    BERT-wwm & 86.1  & 85.2  & 89.4  & 87.0  \\
    BERT-wwm-ext & 86.4  & 85.3  & 89.6  & 87.1  \\
    RoBERTa & 86 & 85 & 89.6 & 87.1 \\
    MacBERT & 86.0  & 85.2  & 89.5  & 87.0  \\
    StyleBERT & \textbf{86.4}  & \textbf{85.3}  & \textbf{90.0}  & \textbf{87.9}  \\
    \bottomrule
    \end{tabular}%
  \setlength{\abovecaptionskip}{0.2cm}  
  \setlength{\belowcaptionskip}{-0.2cm} 
  \caption{Performance of StyleBERT and other models on LCQMC dataset and BQ Corpus.Accuracy is reported for comparison.}
  \label{tab:SPM_result}%
\end{table}%

The goal of SPM is to determine whether two sentences given have the same intent. To test the model, we selected the LCQMC dataset and BQ Corpus for training and evaluation. The LCQMC dataset contains 240K training set, 8.8K validation set and 12.5K test set. Also as a large-scale Chinese dataset, BQ Corpus contains 100K training sets, 10K validation sets and 10K test sets. These datasets are all composed of sentence pairs with ``related'' or ``not related'' label.  \par
From the actual results in Table \ref{tab:SPM_result}, StyleBERT~4 shows the best performance, it surpasses other models.

\subsection{Chinese Named Entity Recognition (Chinese NER)}

\begin{table}[h]
  \centering
    \begin{tabular}{lcccc}
    \toprule
          & \multicolumn{1}{c}{\textit{Weibo}} & \multicolumn{1}{c}{\textit{OntoNotes}} \\
    \textbf{Model} & \multicolumn{1}{c}{\textbf{test-F1}} & \multicolumn{1}{c}{\textbf{test-F1}} \\
    \midrule
    BERT  & 64.4  & 78.2  \\
    BERT-wwm &  69.1  & 78.2  \\
    BERT-wwm-ext & 67.5  & 78.4  \\
    RoBERTa & 66.7 & 79 \\
    MacBERT & 69.3  & \textbf{79.6}  \\
    StyleBERT & \textbf{69.6}  & 79.1  \\
    \bottomrule
    \end{tabular}%
  \setlength{\abovecaptionskip}{0.2cm}  
  \setlength{\belowcaptionskip}{-0.2cm} 
  \caption{Performances of StyleBERT and other models on Weibo
  dataset and Ontonotes dataset.Overall F1 is reported for comparison.}
  \label{tab:NER_result}%
\end{table}%

NER F1 scores can reflect model's ability on token classification tasks. In a NER training dataset, each token is corresponding to a NER label. All named entity are assigned with special NER label indicating their category attribute, like `B-ORG', `B-GPE' or `B-PER'. While other tokens are assigned with a normal label `O'. In this set of experiments, weibo and OntoNotes datasets are selected, which contain social media context. Weibo dataset has small volume of 1.35k training samples, 270 dev samples and 270 test samples. OntoNote contains 15.7k training data, 4.3k dev data and 4.3k test data. As shown in table \ref{tab:NER_result}, StyleBert model out performs all other open source BERT base models by 1-5\%.
\paragraph{}

\section{Conclusion}

In this article, we revisit recently proposed Chinese language models for their major contributions on Chinese character information processing. We also propose the StyleBERT which incorporate word, pinyin, five-stroke and chaizi information for grasp semantics of Chinese characters. And the experiments show that the proposed model is effective in a wide range of Chinese NLP tasks.

\bibliographystyle{unsrtnat}
\bibliography{style_bert}  






\end{document}